\documentclass[conference,fleqn]{IEEEtran}
\usepackage{cite}
\usepackage{amsmath,amssymb,amsfonts}
\usepackage[ruled]{algorithm2e}
\usepackage{graphicx}
\usepackage{textcomp}
\usepackage{xcolor}
\usepackage[utf8]{inputenc}
\usepackage[T1]{fontenc}
\usepackage{booktabs}
\usepackage{multirow}

\def\BibTeX{{\rm B\kern-.05em{\sc i\kern-.025em b}\kern-.08em
    T\kern-.1667em\lower.7ex\hbox{E}\kern-.125emX}}
\begin{document}

\title{Trilevel Memetic Algorithm for the Electric Vehicle Routing Problem}

\author{
\IEEEauthorblockN{1\textsuperscript{st} Ivan Milinović}
\IEEEauthorblockA{\textit{University of Zagreb}\\
\textit{Faculty of Electrical Engineering} \\
\textit{and Computing} \\
ivan.milinovic@fer.hr}
\and
\IEEEauthorblockN{2\textsuperscript{nd} Leon Stjepan Uroić}
\IEEEauthorblockA{\textit{University of Zagreb}\\
\textit{Faculty of Electrical Engineering} \\
\textit{and Computing} \\
leon-stjepan.uroic@fer.hr}
\and
\IEEEauthorblockN{3\textsuperscript{rd} Marko Đurasević}
\IEEEauthorblockA{\textit{University of Zagreb}\\
\textit{Faculty of Electrical Engineering} \\
\textit{and Computing} \\
marko.durasevic@fer.hr}
}

\maketitle

\begin{abstract}
The Electric Vehicle Routing Problem (EVRP) extends the capacitated vehicle routing problem by incorporating battery constraints and charging stations, posing significant optimization challenges. This paper introduces a Trilevel Memetic Algorithm (TMA) that hierarchically optimizes customer sequences, route assignments, and charging station insertions. The method combines genetic algorithms with dynamic programming, ensuring efficient and high-quality solutions. Benchmark tests on WCCI2020 instances show competitive performance, matching best-known results for small-scale cases. While computational demands limit scalability, TMA demonstrates strong potential for sustainable logistics planning.
\end{abstract}

\begin{IEEEkeywords}
electric vehicle routing problem, memetic algorithm, genetic algorithm, vehicle routing, fixed route vehicle charging problem, combinatorial optimization
\end{IEEEkeywords}

\section{Introduction}

The increasing demand for sustainable transportation has positioned electric vehicles (EVs) as a viable alternative to internal combustion engine vehicles. Governments around the world are promoting EV adoption through policy incentives and environmental regulations to mitigate greenhouse gas emissions and reduce dependence on fossil fuels\cite{iea2023}. As a result, electric vehicles are becoming more prevalent in both public and commercial transportation sectors.

Despite their benefits such as zero tailpipe emissions, lower operating costs, and reduced noise pollution, EVs present unique operational challenges. Chief among these are limited driving range, long charging times, and a sparse distribution of charging infrastructure. These factors complicate traditional vehicle routing, especially for logistics operations that require timely and energy-efficient deliveries. Consequently, the Electric Vehicle Routing Problem (EVRP) has emerged as an important variant of the classical Vehicle Routing Problem (VRP), incorporating energy constraints and charging station decisions into the optimization process.

Solving the EVRP is computationally difficult due to its combinatorial nature and the additional layer of complexity introduced by battery management. As exact methods often become impractical for large-scale instances, heuristic and metaheuristic approaches have gained popularity. Among these, Genetic Algorithms (GAs) have proven effective for solving a variety of routing and scheduling problems, thanks to their flexibility, global search capability, and robustness against local optima.

In this study, we propose a GA-based method to solve the EVRP. The main contributions of this paper are described below:
\begin{itemize}
    \item An improved memetic algorithm for finding customer permutations
    \item Trilevel solver structure
    \item Optimal\footnote{For route assignment and charging station insertion as the number of bins approaches infinity.} route assignment and charging station insertion in the optimization loop. 
\end{itemize}

The results demonstrate that genetic algorithms can provide high-quality solutions within reasonable computational times, offering a practical tool for companies transitioning to sustainable transportation models.

Our method improves on the previously proposed solutions based on genetic algorithms and achieves results on par with other state of the art methods.

\section{Problem Statement}

The Electric Vehicle Routing Problem (EVRP) seeks to find optimal routes for a fleet of homogeneous electric vehicles, minimizing the total distance traveled subject to the following constraints: 
\begin{itemize}
    \item Every route must start and end at the depot.
    \item Each customer must be visited exactly once.
    \item The total demand of customers on a given route cannot exceed the vehicle's maximum cargo capacity.
    \item Vehicles must maintain sufficient battery energy throughout the route.
\end{itemize}

Formally, the problem is defined on a weighted, fully connected graph \(G=(V,E)\). 
The vertex set \( V=\{\bar{d}\} \cup V_c \cup \hat{V_s} \) includes a single depot node \(d\), a set of customer nodes \(V_c=\{c_1, c_2, ... ,c_n\}\), and an expanded set of charging stations \( \hat{V_s} \). 
The latter consists of \( \beta_i \) copies of each original charging station \(s_i\) for \( i = 1, 2, ..., m \), ensuring sufficient representation for multiple visits.
The edge set \( E=\{ (i,j), \forall i,j \in V, i \ne j \} \) connects all vertices, with each edge \( (i, j) \) weighted by the Euclidean distance \( d_{ij} \).

When a vehicle traverses edge \( (i,j) \), it consumes \( h\cdot d_{ij} \) units of battery energy, where \(h \in \mathbb{R}^+ \) denotes the battery consumption rate.
All vehicles are identical, sharing the same cargo capacity \(Q\) and battery capacity \(B\).
The EVRP can be formulated mathematically as follows:

\begin{align}
    &min \hspace{1em} f(x) = \sum_{i \in V, j \in V, i \ne j}d_{ij}x_{ij} \label{eq:objective} \quad \quad \quad s.t. \\
    &\sum_{j \in V, i \ne j} x_{ij} = 1, \forall i \in V_c \label{eq:cusomer_visit} \\
    &\sum_{j \in V, i \ne j} x_{ij} \leq 1 \quad \forall i \in \hat{V_s} \label{eq:station_visit} \\
    &\sum_{j \in V} x_{ij} - \sum_{j \in V} x_{ji} = 0 \quad \forall i \in V \label{eq:flow} \\
    &y_j \leq y_i - h d_{ij} x_{ij} + B (1 - x_{ij}) \quad \forall i \in V_c, \forall j \in V, i \ne j \label{eq:battery} \\
    &y_j \leq B - h d_{ij} x_{ij} \quad \forall i \in \hat{V_s} \cup \{d\}, \forall j \in V, i \ne j \label{eq:charging} \\
    &q_j \leq q_i - d_i x_{ij} + Q (1 - x_{ij}) \quad \forall (i,j) \in V, i \ne j \label{eq:cargo} \\
    &x_{ij} \in \{0,1\}, \quad \forall i \in V, \forall j \in V, i \ne j \label{eq:domains}
\end{align}

If the edge \( (i, j) \) is traversed by an electric vehicle, the binary variable \( x_{ij} \) equals one; otherwise, it is set to zero. Equation \ref{eq:objective} defines the EVRP's objective of minimizing the total distance traveled by all EVs. The remaining equations specify the constraints that a feasible solution must satisfy.

Constraint \ref{eq:cusomer_visit} ensures that each customer node is visited exactly once. 
Similarly, Constraint \ref{eq:station_visit} requires that each copy of a charging station is visited at most once. 
Constraint \ref{eq:flow} guarantees flow conservation at every node, meaning the number of incoming edges equals the number of outgoing edges. 
This prevents route discontinuities (i.e., vehicles cannot "teleport" between nodes).
In Constraints \ref{eq:battery} and \ref{eq:charging}, \( y_i \) represents the remaining battery charge level of an EV at node \( i \). 
The former constraint governs battery consumption when traveling between customer nodes, while the latter applies when departing from a charging station or the depot. 
Finally, Constraint~\eqref{eq:cargo} tracks the remaining cargo capacity \( q_i \) at node \( i \), enforcing that the total demand on a route cannot exceed the vehicle's capacity \( Q \) and that the remaining cargo decreases after visiting a customer.

\section{Related Work}
The Electric Vehicle Routing Problem (EVRP) extends the classical capacitated VRP by incorporating electric vehicles limited battery range and the need to visit charging stations. Its first variant was introduced by Erdoğan and Miller-Hooks \cite{erdougan2012green} as the Green Vehicle Routing Problem. Afterwards many electric vehicle specific problem variants came about such as making the time it takes to recharge the vehicle a non-linear function. These extra constraints make EVRP harder than the standard VRP and have motivated numerous studies of EVRP variants (see, e.g., \cite{erdelic2019survey} for a survey). In this work we focus on the basic EVRP without any additional constraints. 

Many exact approaches model EVRP as a 0-1 mixed-integer linear program (MILP). For example, Keskin and Catay \cite{keskin2016partial} extend VRP with time windows (VRPTW) formulations to EVRPTW by including battery energy consumption and allowing partial recharge. Generally, MILP formulations can be solved optimally only on very small problem instances\cite{schneider2014electric}.

Heuristic and metaheuristic methods are prevalent for EVRP\cite{erdelic2019survey}. Ant Colony Optimization (ACO) has been adapted in several studies. For instance, Mavrovouniotis et al.\ \cite{mavrovouniotis2018ant} incorporate a look-ahead feasibility check in ACO so that each EV can always reach a charging station. Jia et al.\ \cite{jia2021ant} propose a bilevel ACO that splits the problem into capacitated VRP (upper level) and fixed route vehicle charging problem (lower level). Similarly, large-neighborhood search methods have been developed. Keskin and Catay \cite{keskin2016partial} design an adaptive large neighborhood search (ALNS) with new destroy-and-repair operators tailored to EV constraints, allowing partial recharging and yielding high-quality solutions. They further combine ALNS with CPLEX to tackle medium-size instances \cite{keskin2018matheuristic}. Schneider et al \cite{schneider2014electric} proposed a hyhrid heuristic that combines VNS with tabu search for EVRPTW. 

Deep reinforcement learning (DRL) has also emerged as a promising approach to tackle this problem. Lin et al.\ \cite{lin2021deep} proposed a DRL framework for the EVRP with time windows, leveraging a hybrid model that integrates graph neural networks and policy gradient methods to learn efficient routing policies. Their results show that DRL can outperform traditional heuristics on medium-scale instances. Building on this line of research, Wang et al.\cite{wang2024end} introduced an end-to-end DRL framework for the EVRP. Their method utilizes a graph attention network (GAT)-based encoder to capture the relational structure of EVRP instances and an attention-based decoder to generate routing sequences. The encoder–decoder architecture, trained via REINFORCE with a baseline, learns a policy network capable of solving large-scale EVRP instances more efficiently than both traditional algorithms and existing DRL baselines.

Evolutionary algorithms, in particular genetic algorithms (GAs), have also been applied to EVRP. For example, Hien et al.\ \cite{hien2023greedy} propose a hybrid Greedy Search + GA (GSGA) where routes are first initialized by a nearest-neighbor clustering (greedy) algorithm and then refined by GA operators. Other GA-based approaches similarly incorporate specialized encoding and repair schemes to enforce battery and recharge feasibility. These evolutionary methods can handle larger instances than exact solvers, but they still face issues like the need to carefully design crossover/mutation operators for the context of electric vehicles. Liu et al\cite{liu2019coordinated} designed a Mixed-Variable Differentiate evolution model for EVCS. Gil-Gala et al. \cite{gil2024evolving} applied Genetic Programming to automatically generate routing policies for EVRP and showed that they perform 20\%-37\% worse to the best known results in negligible time (few miliseconds).

\section{Trilevel Memetic Algorithm}

\subsection{Overview}

In this section, we present the Trilevel Memetic Algorithm (TMA) for solving the Electric Vehicle Routing Problem (EVRP). 
The algorithm decomposes the problem into three hierarchical subproblems, or layers, to enhance computational efficiency and solution quality.

The first layer focuses on generating high-quality customer permutations. 
Solving the combined problem of partitioning customers into routes while simultaneously inserting charging stations would be computationally prohibitive in the optimization loop. 
To address this, we further decompose the problem into two additional subproblems.

The second subproblem takes a customer permutation as input and partitions it into one or more feasible routes by strategically inserting depot visits. 
The third subproblem handles the insertion of charging stations into these routes, which is a well-known problem referred to as the Fixed-Route Electric Vehicle Charging Problem (FR-EVCP).

A key advantage of this approach is that the second and third subproblems can be solved optimally using dynamic programming as the number of bins approaches infinity. While a finite number of bins does not guarantee optimality, it provides a closer approximation to the optimal solution compared to handcrafted heuristic methods.

\begin{figure}[ht]
    \centering
    \includegraphics[width=0.7\linewidth]{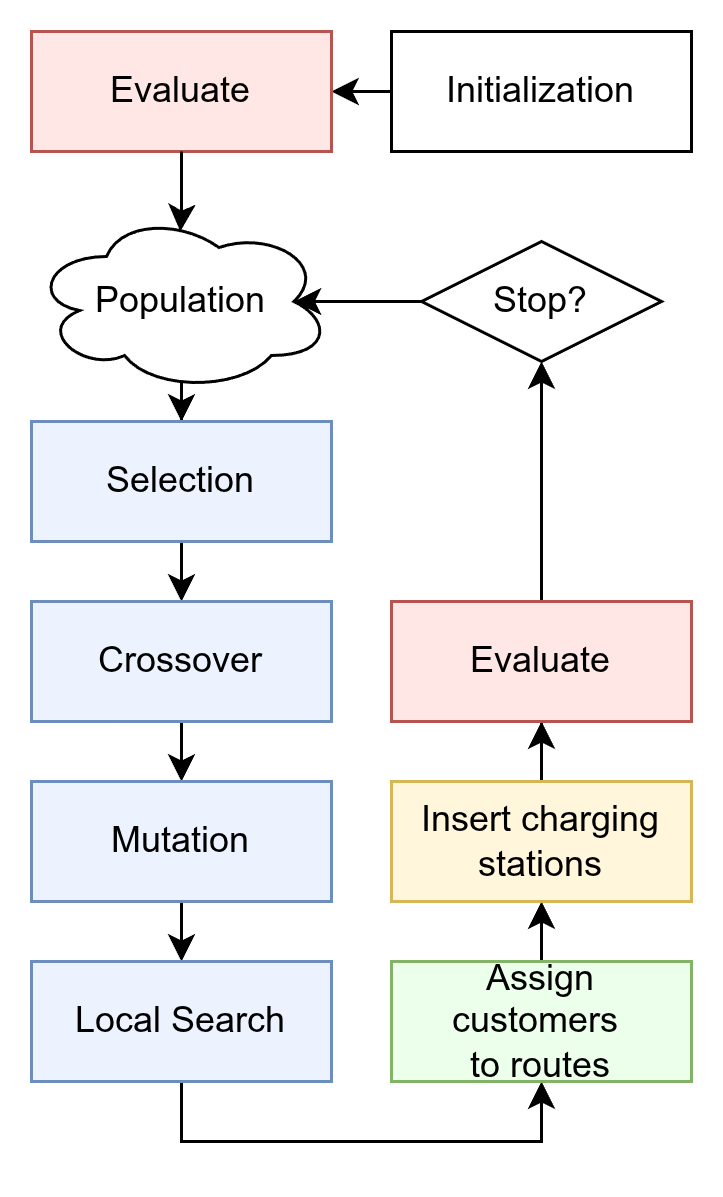}
    \caption{Overview of the trilevel memetic algorithm.}
    \label{fig:algorithm}
\end{figure}

A flowchart illustrating the algorithm's structure is provided in Figure~\ref{fig:algorithm}.

\subsection{Initialization}
The initial population is generated using a stochastic nearest-neighbour algorithm. 
This algorithm constructs customer permutations by iteratively appending customers to the growing sequence. 
At each step, the next customer is selected randomly between $k$ nearest neighbours of the most recently added node in the permutation.

\subsection{Memetic Algorithm}

The memetic algorithm aims to generate high-quality customer permutations, which serve as the foundation for subsequent route optimization. 
Each individual in the population is represented as a permutation of customers only, omitting cargo and charging stations. 
This representation was chosen because route assignments and charging station insertions are optimized in the later stages of the trilevel scheme.

A consequence of this approach is that the fitness (or cost) of an individual can vary depending on the cargo and charging station insertion strategies applied downstream. To ensure consistency, all individuals within a given population must use the same insertion strategies during evaluation.

As an optimization, we enforce population diversity by preventing duplicate solutions. 
This is achieved by storing and comparing solution hashes, though symmetrically equivalent permutations may still coexist.

We used the rank based selection to select parents that are then passed to the following genetic operators in order to create new solutions.

\subsection{Genetic Operators}
Crossover operators are used to explore new regions of the solution space by combining features from parent solutions, promoting diversity and enabling the algorithm to escape local optima.
We used the distributed crossover operator that works in four steps: (1) randomly selecting a customer from both parents; (2) identifying the subroutes (sub1 and sub2) from each parent that contain the selected customer; (3) inheriting remaining customers while preserving their relative order; and (4) constructing two children by concatenating sub2 with sub1 for child1, and the reverse of sub1 with the reverse of sub2 for child2, placing these at the original subroute positions. 

Mutations are used to introduce small changes in individual solutions, improving exploration.
Two different mutation operators are used: heuristic-swap mutation which takes a random customer and swaps it with the closest customer from a different route and heuristic-move mutation which takes a random customer and moves the closest customer from a different route to the route of the first customer. For more detailed explanation of the genetic operators look at the paper from Hien. et al\cite{hien2023greedy}. 

\subsection{Local Search}

\begin{algorithm}[ht]
\caption{Local Search Algorithm}\label{alg:local_search}
\SetAlgoLined
\KwIn{Customer permutation $P = (c_1, c_2, ..., c_n), L \in \{2Opt, Swap\}$}
\KwOut{Improved customer permutation $P'$}
\SetKwFunction{LocalSearch}{LocalSearch}
\SetKwProg{Fn}{Function}{}{}

\Fn{\LocalSearch{$P$}}{
    $\text{improvement} \gets \text{True}$\;
    \While{improvement}{
        $\text{improvement} \gets \text{False}$\;
        \For{Each route spanning from $c_k$ to $c_l$ }{
            \For{$i \gets l$ \KwTo $k-1$}{
                \For{$j \gets i+1$ \KwTo $l$}{
                    \If{the distance after applying $L$ is smaller}{
                        Apply $L$ to the segment $c_k$, $c_l$\;
                        $\text{improvement} \gets \text{True}$\;
                    }
                }
            }
        }
    }
    \Return{$P$}\;
}
\end{algorithm}

The local search procedure improves the order of customers optimizing the intra-route customer sequence by minimizing inter-customer distances. 
This design choice deliberately excludes depot and charging station considerations to maintain computational efficiency, as reevaluating the complete solution after each local move would be prohibitively expensive.

Two neighbourhood operators are employed. The 2-opt operator, which improves routes by removing two non-adjacent edges and reconnecting the reversed intermediate segment and the swap operator exchanges positions of customer pairs within the same route. 
The algorithm systematically explores these neighborhoods until no further improving moves can be found, as detailed in Algorithm~\ref{alg:local_search}. This focused optimization at the customer permutation level provides a computationally efficient mechanism for route improvement before subsequent stages handle depot and charging station insertions.

\subsection{Route assignment}

The route assignment algorithm takes a customer permutation as input and partitions it into one or more feasible routes by strategically inserting depot visits. 
The algorithm preserves the original customer visitation order while ensuring all cargo capacity constraints are satisfied. 
Its objective is to minimize total travel distance without considering energy constraints.

The route assignment is a special case of the constrained shortest path problem which is known to be NP-hard even with a single constraint~\cite{hartmanis1982computers}.
To solve this problem we employed a dynamic programming approach which has a pseudo-polynomial complexity.
Fortunately, all customer demands and vehicle cargo capacities are integers so we were able to obtain an optimal solution for every instance and every permutation of customers in reasonable time.

A cell in a dynamic programming table \( dp[i][j] \) stores the minimal distance in which a vehicle can reach first \(j\) customers in the permutation while still having \(i\) units of cargo capacity left after visiting the \(j\)-th customer.
All cells of the table are initially set to infinity except the cell \( dp[Q][0] = 0 \) because we assume that the \(j=0\) and \(j=n\) nodes are depot nodes because we need to have at  least a single route.
We than iterate over \(j=0,1,2...n\) and fill the dp table.
At each step we can go directly from the \((j-1)\)-th customer in the permutation to the \(j\)th customer according to the equation \ref{eq:cargo_dp_step1} in which \(d_j\) denotes the demand of the \(j\)th customer in the permutation.
On the other hand, we could visit the depot and update the table according to the equation \ref{eq:cargo_dp_step2}.

\begin{equation}
    \label{eq:cargo_dp_step1}
    \begin{split}
        dp[i][j] = min(& dp[i][j], \\
               & dp[i+d_j][j-1] + \\
               & distance(c_{j-1}, c_{j}))
    \end{split}
\end{equation}

\begin{equation}
    \label{eq:cargo_dp_step2}
    \begin{split}
        dp[Q-d_j-1][j] = min(& dp[Q-d_j-1][j], \\
                     & dp[i][j-1] + \\
                     & distance(c_{j-1}, \bar{d}) + \\
                     & distance(\bar{d}, c_{j}))
    \end{split}
\end{equation}

After the table is filled, we find the smallest distance in the last column and backtrack the solution through the table.

\subsection{Charging station insertion}

The third-level problem is a well-known Fixed-Route Electric Vehicle Charging Problem (FR-EVCP), which is a special case of the Constrained Shortest Path problem and thus NP-hard. 
Our method extends the approach of Deschênes et al.~\cite{deschenes2022dynamic} by allowing visits to multiple charging stations between customer nodes. 

Since the distances (and thus energy requirements) between nodes are real-valued, we discretize the battery capacity into \( K \) bins to enable dynamic programming. 
While this discretization yields a suboptimal solution, the approximation converges to the optimal solution as \( K \) increases.

Similar to the route assignment algorithm, we use a dynamic programming table \( dp[i][j] \) that stores the minimum distance required to serve the first \( j \) customers with \( i \) units of residual energy at the \( j \)-th node. The table is filled iteratively for \( j = 0, 1, \dots, n \). At each step, we consider whether to charge the vehicle or not.

For the case where we don't charge, the DP table is updated according to Equation~\ref{eq:energy_dp_step1}. 
If we decide to charge the vehicle, the DP table is updated according to Equation~\ref{eq:energy_dp_step2}. 
In latter equation, we consider every possible combination of entry charging station \(encs\) and exit charging station \(excs\). 
The entry charging station is important because we must be able to reach it with the remaining energy, while the exit charging station determines how much energy will remain when arriving at the \(j\)-th customer in the permutation.
The charging stations visited between $encs$ and $excs$ are independent of the dynamic programming variables, allowing us to precompute the optimal path between them using an A* algorithm and store the results in table $C$.

\begin{equation}
    \label{eq:energy_dp_step1}
    \begin{split}
        dp[i][j] = min(& dp[i][j], \\
                       & dp[i+e_{j,j-1}][j - 1] +\\
                       & distance(c_{j-1}, c_{j}))
    \end{split}
\end{equation}

\begin{equation}
    \label{eq:energy_dp_step2}
    \begin{split}
        dp[K-&e_{encs, j-1} - 1][j] = min(\\
                                        &dp[K-e_{ecs, j-1} - 1][j], \\
                                        & dp[i][j - 1] + distance(c_{j-1}, encs) + \\
                                        & C[encs][excs] + distance(excs, c_j) \\
                                        ) \quad &
    \end{split}
\end{equation}

The time complexity of the proposed method is \(\mathcal{O}(nKm^2)\), where \(n\) denotes the number of customer nodes, \(m\) represents the number of charging stations, and \(K\) corresponds to the number of discretization bins. Due to this high complexity, the algorithm faces computational challenges when applied to large-scale instances or instances with many charging station.

\section{Experiments}

\subsection{Dataset}

To evaluate the performance of our proposed method, we employed the benchmark dataset from the IEEE WCCI2020 competition \cite{mavrovouniotis2020benchmark}. 
The dataset consists of two distinct subsets designed to test algorithmic performance across different scales. The first subset comprises seven small-scale instances with up to 101 customers, derived from the classical vehicle routing benchmarks introduced by Christofides and Eilon~\cite{christofides1969algorithm}. 
The second subset contains large-scale instances extended from the Uchoa et~al. dataset~\cite{uchoa2017new}, featuring up to 1,001 customer nodes. 

\subsection{Experimental Setup}
Our experimental protocol followed the evaluation framework of the WCCI~2020 competition~\cite{mavrovouniotis2020benchmark}. 
For each problem instance, we executed 20 independent runs with different random seeds to account for stochastic variability in the results. 
Each run was limited to a maximum of \(25,\!000n\) objective function evaluations, where \(n = |V_c| + |V_s| + 1\) represents the total number of nodes including customers, charging stations, and the depot.
All experiments were conducted with the same hyperparameters listed in table~\ref{tb:hyperparameters}.
The number of generations was adjusted for each instance to comply with the previously mentioned restriction on the number of evaluations. 

After the algorithm returned a solution, we further optimized the charging station placements by running the energy repair procedure with a high number of bins (100,001 as specified in Table~\ref{tb:hyperparameters}).

Additionally, a few problem instances with a lot of charging stations were excluded from the evaluation due to slow performance of the charging station insertion procedure which made them impractical to run under the same experimental conditions with a significant number of generations.

\subsection{Results}

The experimental results, as presented in Table~\ref{tb:results}, reveal the competitive performance of the proposed Trilevel Memetic Algorithm (TMA) in solving small-scale instances of the EVRP benchmark set. TMA consistently matches the best-known solutions for instances E22, E23, and E30, achieving zero standard deviation.

In more challenging instances like E33, E51, E76, and E101, TMA yields solutions that are generally close to the best-known results but with slightly higher variance. For example, in instance E33, the best TMA solution is equal to the best solutions of most competing methods, but the mean performance is marginally worse. Similarly, for E76 and E101, the TMA displays larger standard deviations, indicating potential room for improvement in the fine-tuning of the hyperparameters.

TMA matches or surpasses the performance of GA in terms of best solutions for most instances. However, the Bilevel Ant Colony Optimization (BACO) algorithm shows slightly better performance in most cases. 

It is important to note that, due to the relatively slow computation of the charging station insertion component, TMA was run for substantially fewer iterations than the maximum allowed by the benchmark rules. This computational bottleneck limited the extent of the search process, especially in larger instances where evaluating charging station feasibility becomes more complex. Despite this constraint, TMA still achieves competitive best solutions, suggesting that with more runtime or a more efficient insertion routine, overall performance could improve significantly.

\begin{table}[ht]
\centering
\caption{Hyperparameters}
\label{tb:hyperparameters}
\begin{tabular}{lc}
\toprule
\textbf{Hyperparameter} & \textbf{Value} \\
\midrule
Stochastic nearest neighbour candidates & 3 \\
Population size & 200 \\
Selection pressure & 1.6 \\
Number of elite individuals & 30 \\
Energy repair bin count while opt. & 151 \\
Finishing energy repair bin count & 100,001 \\
\bottomrule
\end{tabular}
\end{table}

\subsection{Comparative Analysis} \label{subsec:comparison}
We evaluated the performance of our proposed method against five state-of-the-art algorithms for EVRP. 
Three top-performing approaches from the WCCI~2020 competition were included: a genetic algorithm (GA), simulated annealing (SA), and variable neighbourhood search (VNS). 
Additionally, we implemented the iterated local search (ILS) method originally developed by Liang et al. \cite{liang2021electric} for EVRP with non-linear charging characteristics, which we adapted to the standard EVRP formulation through minor modifications. 
Finally, we incorporated the Bilevel Ant Colony Optimization (BACO) algorithm proposed by Jia et al. \cite{jia2021ant}.

\begin{table*}[ht]
\centering
\caption{Experimental Results}
\label{tb:results}
\begin{tabular}{lccccccc|c}
\toprule
Case & Index & BKS & ILS & GA & SA & VNS & BACO & \textbf{TMA} \\
\midrule
\multirow{3}{*}{E22} & min & 384.67 & 384.67 & 384.67 & 384.67 & 384.67 & 384.67 & 384.67 \\
 & mean & & 385.69 & 384.67 & 384.67 & 384.67 & 384.67 & 384.67 \\
 & std & & 2.11 & 0.00 & 0.00 & 0.00 & 0.00 & 0.00 \\
\addlinespace
\multirow{3}{*}{E23} & min & 571.94 & 590.35 & 571.94 & 571.94 & 571.94 & 571.94 & 571.94 \\
 & mean & & 592.05 & 571.94 & 571.94 & 571.94 & 571.94 & 571.94 \\
 & std & & 1.04 & 0.00 & 0.00 & 0.00 & 0.00 & 0.00 \\
\addlinespace
\multirow{3}{*}{E30} & min & 509.47 & 509.47 & 509.47 & 509.47 & 509.47 & 509.47 & 509.47\\
 & mean & & 509.47 & 509.47 & 509.47 & 509.47 & 509.47 & 509.47\\
 & std & & 0.00 & 0.00 & 0.00 & 0.00 & 0.00 & 0.00 \\
\addlinespace
\multirow{3}{*}{E33} & min & 840.14 & 840.57 & 844.25 & 840.57 & 840.14 & 840.57 & 840.57 \\
 & mean & & 844.07 & 845.62 & 854.07 & 840.43 & 842.30 & 842.98 \\
 & std & & 7.78 & 0.92 & 12.80 & 1.18 & 1.42 & 2.17 \\
\addlinespace
\multirow{3}{*}{E51} & min & 529.90 & 529.90 & 529.90 & 533.66 & 529.90 & 529.90 & 529.90 \\
 & mean & & 539.03 & 542.08 & 533.66 & 543.26 & 529.90 & 541.39 \\
 & std & & 6.92 & 8.57 & 0.00 & 3.52 & 0.00 & 11.43 \\
\addlinespace
\multirow{3}{*}{E76} & min & 692.64 & 694.64 & 697.27 & 701.03 & 692.64 & 692.64 & 696.80 \\
 & mean & & 704.24 & 717.30 & 712.17 & 697.89 & 692.85 & 712.54 \\
 & std & & 7.15 & 9.58 & 5.78 & 3.09 & 0.81 & 9.90 \\
\addlinespace
\multirow{3}{*}{E101} & min & 839.29 & 841.02 & 852.69 & 845.84 & 839.29 & 840.25 & 845.07 \\
 & mean & & 851.62 & 872.69 & 852.48 & 853.34 & 845.95 & 864.76 \\
 & std & & 6.93 & 9.58 & 3.44 & 4.73 & 4.58 & 9.36 \\
\addlinespace
\multirow{3}{*}{X143} & min & 15901.23 & 16058.29 & 16488.60 & 16610.37 & 16028.05 & 15901.23 & 16183.2 \\
 & mean & & 16318.57 & 16911.50 & 17188.90 & 16459.31 & 16031.46 & 16627.40 \\
 & std & & 160.07 & 282.30 & 170.44 & 242.59 & 262.47 & 293.40 \\
\addlinespace
\multirow{3}{*}{X214} & min & 11133.14 & 11323.56 & 11762.07 & 11404.44 & 11323.56 & 11133.14 & 11721 \\
 & mean & & 11537.58 & 12007.06 & 11680.35 & 11482.20 & 11219.70 & 121842 \\
 & std & & 72.55 & 156.69 & 116.47 & 76.14 & 46.25 & 169.42 \\
\addlinespace
\multirow{3}{*}{X573} & min & 51929.24 & 53102.46 & 54189.62 & 51929.24 & 52181.51 & 53822.87 & 53266.58 \\
 & mean & & 53507.46 & 55327.62 & 52793.66& 52548.09 & 54567.15 & 53587.75 \\
 & std & & 275.76 & 548.05 & 577.24 & 278.85 & 231.05 & 485.97 \\
\addlinespace
\multirow{3}{*}{X916} & min & 341649.91 & 348733.86 & 357391.57 & 342796.88 & 341649.91 & 342993.01 & 350161.57 \\
 & mean & & 350822.41 & 360269.94 & 343533.85& 342460.70 & 344999.95 & 350982.45 \\
 & std & & 1177.08 & 229.19 & 556.98 & 510.66 & 905.72 & 649.60 \\
\addlinespace
\multirow{3}{*}{X1001} & min & 76297.09 & 79493.37 & 78832.90 & 78053.86 & 77476.36 & 76297.09 & 80044.70 \\
 & mean & & 79928.29 & 79163.34 & NA & 77920.52 & 77434.33 & 81358.13 \\
 & std & & 265.91 & NA & 306.27 & 234.73 & 719.87 & 737.44 \\
\bottomrule
\end{tabular}
\end{table*}

\section{Conclusion}

In this work, we evaluated the proposed trilevel memetic algorithm within the standardized framework of the WCCI~2020 competition, ensuring consistency and comparability with prior studies. While some problem instances had to be excluded due to computational bottlenecks in the charging station insertion procedure, our results across the remaining instances demonstrate that the method performs robustly even with a limited number of generations. This highlights the potential of our approach and motivates future work on improving scalability and efficiency for larger or more complex instances.

\bibliographystyle{IEEEtran}
\bibliography{references}

\end{document}